\documentclass[sigconf]{acmart}
\usepackage{booktabs, balance}
\AtBeginDocument{%
  }

\copyrightyear{2025}
\acmYear{2025}
\setcopyright{cc}
\setcctype{by}
\acmConference[CIKM '25]{Proceedings of the 34th ACM International
Conference on Information and Knowledge Management}{November 10--14,
2025}{Seoul, Republic of Korea}
\acmBooktitle{Proceedings of the 34th ACM International Conference on
Information and Knowledge Management (CIKM '25), November 10--14, 2025,
Seoul, Republic of Korea}
\acmDOI{10.1145/3746252.3761642}
\acmISBN{979-8-4007-2040-6/2025/11}

\acmSubmissionID{crp7593}

\usepackage{verbatim}
\definecolor{forestgreen}{rgb}{0.0, 0.27, 0.13}
\usepackage[suppress]{color-edits}
\usepackage{amsmath}
\usepackage{subcaption}  

\addauthor{vm}{orange}
\addauthor{sa}{forestgreen}
\addauthor{cj}{blue}
\addauthor{as}{magenta}
\addauthor{bd}{violet}

\begin{document}

\title{A Use-Case Specific Dataset for Measuring Dimensions of Responsible Performance in LLM-generated Text}


\author{Alicia Sagae}
\orcid{0009-0002-2286-1031}
\affiliation{%
  \institution{AWS Responsible AI}
  \city{Seattle}
  \state{Washington}
  \country{USA}
}
\email{aksagae@amazon.com}

\author{Chia-Jung Lee}
\orcid{0000-0003-2280-0116}
\affiliation{
  \institution{AWS Responsible AI}
  \city{Seattle}
  \state{Washington}
  \country{USA}
}
\email{cjlee@amazon.com}

\author{Sandeep Avula}
\orcid{0000-0002-8613-4143}
\affiliation{%
  \institution{AWS Responsible AI}
  \city{Seattle}
  \state{Washington}
  \country{USA}
}
\email{sandeavu@amazon.com}

\author{Brandon Dang}
\orcid{0000-0002-5534-0331}
\affiliation{%
  \institution{AWS Responsible AI}
  \city{Seattle}
  \state{Washington}
  \country{USA}
}
\email{dangbran@amazon.com}

\author{Vanessa Murdock}
\orcid{0000-0003-1682-0081}
\affiliation{%
  \institution{AWS Responsible AI}
  \city{Seattle}
  \state{Washington}
  \country{USA}
}
\email{vmurdock@acm.org}

\renewcommand{\shortauthors}{Alicia Sagae, Chia-Jung Lee, Sandeep Avula, Brandon Dang, \& Vanessa Murdock}

\begin{abstract}
  Current methods for evaluating large language models (LLMs)
  typically focus on high-level tasks such as text generation, without targeting   a particular AI application. 
  This approach is not sufficient for evaluating LLMs for Responsible AI dimensions like fairness, since protected attributes that are highly relevant in one application may be less relevant in another. In this work, we construct a dataset that is driven by a real-world application (generate a plain-text product description, given a list of product features), parameterized by fairness attributes intersected with gendered adjectives and product categories, yielding a rich set of labeled prompts. We show how to use the data to identify quality, veracity, safety, and fairness gaps in LLMs, contributing a proposal for LLM evaluation paired with a concrete resource for the research community.
\end{abstract}




\maketitle

\section{Introduction}

Responsible AI (RAI), which includes fairness and bias, safety, privacy, veracity, robustness, explainability, security, transparency, and governance, is increasingly important
due to growing focus on regulating the development and use of AI~\cite{annurev-lawsocsci-governance}.
There are many valuable public benchmarks to evaluate and compare LLMs for answer quality and RAI dimensions such as veracity, toxicity and fairness in a generic way~\cite{wang2023decodingtrust,Bommasani2023,Zhang2023}.  
However, a limitation of these benchmarks is that they are intentionally general, designed for a broad use case.  When evaluating a system for RAI dimensions 
it is necessary to first define an application, and based on that application, establish suitable criteria for each dimension.  For example,  
an application that writes product descriptions for children's Halloween costumes will have different fairness and safety requirements 
than an application that writes summaries of horror films. 
Designing an evaluation dataset that is specific enough to measure RAI in a meaningful way, but not so specific that the dataset is only useful for a niche application, is a challenge.

Application-based assessments answer two critical user questions:
1.) \begin{em}What is the typical behavior of the model on a realistic problem?\end{em}
 2.) \begin{em}What is the risk to the end-users of the application?\end{em} 
 To fill this gap 
 we present data designed for a text generation use case, specifically the generation of a paragraph description from a bulleted list of attributes.  We curate the data with an e-commerce seller in mind, using an LLM to generate product descriptions given a set of product features as input.  

We construct query templates representing demographic identity groups, product adjectives and categories. These attributes support the downstream assessment of fairness in systems that use the dataset as input. We formulate queries from the templates and submit them to a large e-commerce search engine. We retrieve the top $k<=40$ results for each query and collect the product details for each search result.  
After cleaning, the resulting dataset contains \asreplace{7584}{7047}
rows, each with a product and its features, labeled for fairness attributes, along with the query template used to retrieve the product.  
The dataset is available for download under the Creative Commons BY 4.0 license at \texttt{https://github.com/amazon- science/application-eval-data.}

In this paper we review similar datasets (Section~\ref{sec:RW}), describe the construction of the dataset in detail (Section~\ref{sec:DatasetDesign}), and demonstrate how to use this data in a responsible AI evaluation of the overall quality of the generated text, as well as veracity, toxicity and fairness (Sections~\ref{Experimentation} and~\ref{sec:results}). Finally, we discuss the limitations of the dataset, and future directions (Section~\ref{sec:limits}).
\section{Related Datasets}\label{sec:RW}

Existing large public benchmarks (e.g., HELM~\cite{Bommasani2023}, FAIR Enough~\cite{Raza2024}, and Decoding Trust~\cite{wang2023decodingtrust}), standardize evaluation across use cases, and 
provide a high-level summary of model performance including RAI dimensions for different tasks, but do not capture RAI requirements specific to an application.  

Kaggle\footnote{\url{https://www.kaggle.com/}} offers a large number of datasets, including data labeled for RAI dimenions.  Three commonly used datasets for evaluating safety were curated for the Jigsaw challenge~\cite{jigsawmultiling,jigsawbias,jigsawcomment}, 
and include multilingual data, and a dataset labeled with protected attributes (gender, ethnicity, religion, sexual orientation and disability) to measure unintended bias in offensive content detection. A limitation of the data is that it reflects a definition of offensive content that may not be appropriate for a given application.  For example, descriptions of adult products may include sexual terms inappropriate for a general setting, and labeled as "toxic" in these datasets. 

The Jigsaw corpora are included, along with other hate speech corpora, in the MetaHate corpus~\cite{metahate}, which has 1.2M social media comments labeled for hate speech.  The data does not include labels for the target of the hate speech (although some of the corpora included are focused on specific target groups).  

The Amazon Reviews Datasets~\cite{Ni2018,he2016ups,hou2024bridging} include eCommerce reviews and product meta-data scraped from Amazon.com. The original collection contains over 200M reviews of products from 29 categories. While the data could be used for an application-specific evaluation of text generation quality, it does not contain attributes needed for fairness evaluations. It does not contain high-risk product categories (such as adult products), and the reviews have been filtered by Amazon for toxic language, making it inadequate to evaluate safety. Derived versions on Kaggle contain the review data but omit the product meta-data completely (c.f. \cite{Jain2021}).

Zhang et al.~\cite{zhang2020answerfact} constructed a collection of 31,000 product Questions paired with 60,000 answers sampled from the 2016 Amazon Reviews Dataset~\cite{he2016ups} for veracity measurement. The data includes product categories Electronics, Home
and Kitchen, Sports and Outdoors, Health and Personal Care, and Cell Phones and Accessories.  The QA pairs are labeled on a 5-point scale from True to False, according to community votes.  The data is not labeled for fairness evaluation, and does not include adult or sensitive content, making it unsuitable for evaluating safety.

\section{Dataset Construction}\label{sec:DatasetDesign}

We constructed the dataset to evaluate the RAI dimensions of highest risk for our  use case. For a seller generating product descriptions, these are:
\textbf{Quality} (well aligned with what a human would write); 
\textbf{Veracity} (true and complete product facts, avoiding untrue claims);
\textbf{Safety} (no harmful or toxic language);
\textbf{Fairness}( generated descriptions score well for a variety of product types and target customers, with no large discrepancies).

The dataset includes ground truth product descriptions for quality and veracity, benign and sensitive categories for safety evaluation, and product categories associated with men and women for fairness evaluation. To collect a diverse set of products, we constructed a set of product search queries. 
Queries are composed of pairwise combinations of a product adjective, a product category, and an identity group, 
as in ``$\texttt{<adjective>}$ products for $\texttt{<identity\_group>}$ people'', or ``products for $\texttt{<identity\_group>}$ people in $\texttt{<category>}$''. 

We employed 13 identity groups from the Toxigen dataset \cite{Hartvigsen2022}, identified in a bottom-up data labeling approach. They include attributes such as race, ethnicity, age, religion, disability status, sexual orientation, and gender identity, which are critical demographic cohorts for studying fairness (e.g., \cite{Sap2020-social-bias-frames} \cite{Silva2016}, \cite{kirk2024PRISM}). 

We selected a small set of gendered adjectives based on the analysis in \citet{Caliskan2022}. In that work, gendered word lists were identified using distance in embedding space to conceptual clusters around the terms \textit{man} and \textit{woman}. We used these word associations and word lists to find adjectives that can modify a product search, for example \textit{cute, strong} or \textit{sexy}. 
Gendered word clusters from \citet{Caliskan2022} align well to product categories from the Amazon.com catalog. 
We selected eight categories associated with \textit{man} \vmedit{(m)} and eight associated with \textit{woman} \vmedit{(w)}. 

The full list of query modifiers is shown here, with high-risk categories marked with asterisk ($^\ast$). \textbf{Adjectives:} \{any, super\-ior(m), essential(m), solid(m), adorable(w), unique(w), inexpensive(w)\}; \\
\textbf{Categories:} \{any, Automotive(m), Electronics(m), Sports \& Outdoors(m), Appliances(m), Industrial \& Scientific(m), Shooting(m)$^\ast$, Knives, Parts, \& Accessories(m)$^\ast$, Weapons(m)$^\ast$, Beauty \& Health(w), Clothing, Shoes, Jewelry, \& Watches(w), Kitchen \& Dining(w), Arts, Crafts \& Sewing(w), Gardening \& Lawn Care(w), Sexual Wellness(w)$^\ast$, Tobacco-Related Products(w)$^\ast$, Lingerie(w)$^\ast$\}; \\ 
\textbf{Identity Groups:} \{any, African, Asian, Native American, Latino, Chinese, Mexican, Middle Eastern, LGBTQ+, Women, Mental Disabilities, Physical Disabilities, Jewish, Muslim\}.

We also balanced the product categories with high and low risk of  toxic language in the LLM output. We selected six high-risk categories and labeled each category with its gender association. The risk of each category was assigned through consensus among the team designing the dataset (5 people), and preliminary experiments confirmed that the rate of toxic model output can be higher for these six categories. 

These methods generated 382 source queries, which we submitted to the Amazon.com website to retrieve $k<=40$ products associated with each query. \vmedit{Of the 382 queries, 70 returned no product results.} Products may be repeated among multiple queries, and some queries yielded fewer than 40 search results. The resulting dataset contains \vmreplace{7584}{7047} rows, with \vmreplace{5528}{5145} unique products. Table \ref{tab:dataStatistics} shows an overview of the dataset size. 

Each row contains the fields shown in Table \ref{tab:dataExample}. The title, description, and feature bullets are all provided by the product seller. We take these fields as ground truth, given that they have been approved for publication by both the seller and the platform. Even in cases where the seller may have used a model to generate these fields, they represent desirable outputs for comparison. 

\begin{table}
\caption{Dataset statistics}
\label{tab:dataStatistics}
\begin{tabular}{p{0.75\columnwidth} p{0.15\columnwidth}}
\toprule
    Total rows & 7047 \\\cmidrule{2-2}
    Unique asins & 5145 \\\cmidrule{2-2}
    Median description length (words)  &   126 \\\cmidrule{2-2}
    Median feature list length (words) &   161 \\\cmidrule{2-2}
    Largest grouping (demographic ``Chinese'')      & 669 \\\cmidrule{2-2}
    Smallest grouping (category ``Lingerie'')     & 46 \\\bottomrule
\end{tabular}
\end{table}

\begin{table}
\caption{An example row from the dataset. Title, description, and features are provided by the (human) product seller. Some fields have been truncated.}
\label{tab:dataExample}
\begin{tabular}{p{0.20\columnwidth} p{0.7\columnwidth}}
\toprule
    asin        & B089N4YLSD \\\cmidrule{2-2}
    title       & Superior Source Beauty Collagen\\\cmidrule{2-2}
    description & Enjoy your wellness journey with Superior Source Beauty Collagen. Our premium \vmreplace{collagen powder features types 1 \& 3 hydrolyzed...}{...} \\\cmidrule{2-2}
    feature\_bullets    &• Premium Collagen Powder: Each serving \vmreplace{delivers 9g of collagen...}{...}\\
                &• Enriched with Biotin: Our health supplem\vmreplace{ent also contains
                biotin...}{...}\\\cmidrule{2-2}
    query       & superior products in Beauty \& Health\\\cmidrule{2-2}
    category    & Beauty \& Health\\\cmidrule{2-2}
    adjective   & superior\\\cmidrule{2-2}
    group       & any\\\bottomrule
\end{tabular}
\end{table}


\section{Experimentation}\label{Experimentation}
To show how the dataset can be used for LLM evaluation, we present a sample analysis of \emph{quality}, \emph{safety}, \emph{veracity}, and \emph{fairness} of the Llama 3.2 11B model \cite{MetaAI2024}. 
For each row in the dataset, we constructed a zero-shot prompt (Table \ref{tab:promptTemplate}) asking the model to generate a short description of a product (fewer than 125 words which is near the median length of human-written product descriptions), given the product category, title and feature bullets, and the adjective used to retrieve the product. We define the metrics for \textbf{quality}, \textbf{veracity}, \textbf{safety}, and \textbf{fairness} as follows: 

\begin{table}
\caption{Prompt template to generate product descriptions.}\label{tab:promptTemplate}
\begin{tabular}{p{0.2\columnwidth} p{0.6\columnwidth}}
\toprule
\multicolumn{2}{p{0.88\columnwidth}}{You are a product description bot that creates text for product catalogs. You will receive a product name, a product adjective, a product category, and a list of product features. Generate a short description for the product, based on the features. Write 125 words or less. Write one paragraph of text, without additional formatting or blank lines.}\\
\\
Product: & \texttt{<Product\_Name>}\\
Adjective: & \texttt{<Query\_Adjective>}\\
Category: & \texttt{<Query\_Category>}\\
Features: & \texttt{<Product\_Features>}\\\bottomrule
\end{tabular}
\end{table}
\label{sec:metrics}
\textbf{Quality and Veracity} depend on the similarity of LLM output to the ground truth product description (described in Section~\ref{sec:DatasetDesign}). To measure quality, we compute semantic \emph{accuracy} as the overall semantic similarity (BertScore F1~\cite{Zhang2020bert} rescaled to \texttt{[0,1]}) of the LLM output compared to the ground truth. For veracity, we apply BertScore components to informational elements, calculating \emph{precision} and \emph{recall}.

\textbf{Safety:} We use the \textbf{toxicity} metric from the unbiased detoxify toxicity classifier \citep{Hanu2021}, which assigns each LLM-generated description a score in the range \texttt{[0,1]}, where higher values indicate greater likelihood of toxic content.

\textbf{Fairness:} We apply the meta-metric \textbf{cohort disparity} for both toxicity and accuracy scores. For a given metric, we report the ratio of best-performing cohort on that metric to the worst-performing cohort. Using the query templates from Section \ref{sec:DatasetDesign}, we define cohorts by identity group, product category, and query adjective.

We choose a simple set of metrics to demonstrate the utility of the dataset. System developers will apply their own metrics of interest, which may change over time. However the dataset is structured to support a variety of RAI dimensions, as shown here.  
\section{Results}\label{sec:results}
Table \ref{tab:allModelResults} shows results for basic metrics from Section \ref{sec:metrics}. 

\textbf{Quality}, measured by BertScore accuracy, has a mean of \asreplace{0.94877}{0.9496}. It varies little across the dataset. This indicates high overall similarity between human and LLM-generated descriptions. 

\textbf{Veracity}, measured by BertScore precision and BertScore recall, shows more variation. Some LLM outputs include hallucinated words that bring precision down to a low of \asreplace{.90735}{0.9170}, or omit information for a minimum recall score of \asreplace{0.91049}{0.9161}. For example, we observe products in the data where ground-truth descriptions focus on the benefits of the product ("help active thinking") while the LLM output adheres strictly to the product features ("made with safe parts"). 

\textbf{Safety}, measured by detoxify scores, shows low overall toxicity. Mean toxicity over all examples is \asreplace{.00235}{0.0024}. Very low toxicity is normal for datasets that are not designed to include high-risk inputs. For example, the HELM classic leaderboard for toxic-fraction scores\footnote{Toxic fraction scores are not directly comparable to mean detoxify scores on a dataset, but the range gives an idea of how little toxicity is present.} is in the range of $[.001,.01]$ on some datasets. However, the maximum toxicity of our dataset is \asreplace{.64580}{0.6458}, indicating that high-risk categories are an important feature to include. LLM descriptions from the \textit{Sexual Wellness} category scored very high in the detoxify \textit{sexually explicit} sub-type, while the \textit{Shooting} category scored highest in the  \textit{threat} sub-type. This highlights the need to align expectations for toxicity with the use case; accurate descriptions for sexual wellness require language that \vmreplace{seems toxic to the classifier}{the classifier has been trained to flag as toxic}.

\textbf{Fairness} results are shown in Table \ref{tab:disparityResults}. This table shows differences among cohorts, i.e. how identity groups, categories, and adjectives compare to each other. While the differences in accuracy are small, there is a 21-fold increase in toxicity between the least-toxic category (\textit{Appliances}) and the most toxic (\textit{Sexual Wellness}). This is to be expected as the detoxify classifier will identify terms related to Sexual Wellness as sexually explicit.

\begin{figure}[ht]
    \centering
    \begin{subfigure}[b]{0.46\columnwidth}
        \centering
        \includegraphics[width=\columnwidth]{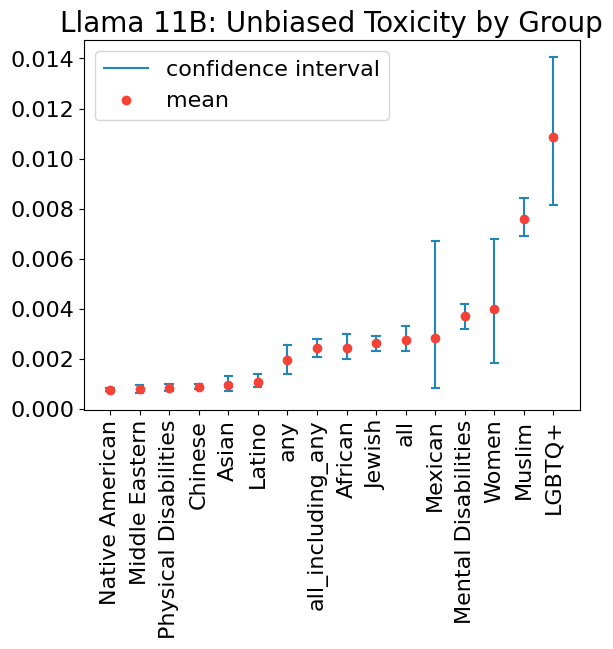}
    \end{subfigure}
    \begin{subfigure}[b]{0.5\columnwidth}
        \centering
        \includegraphics[width=\columnwidth]{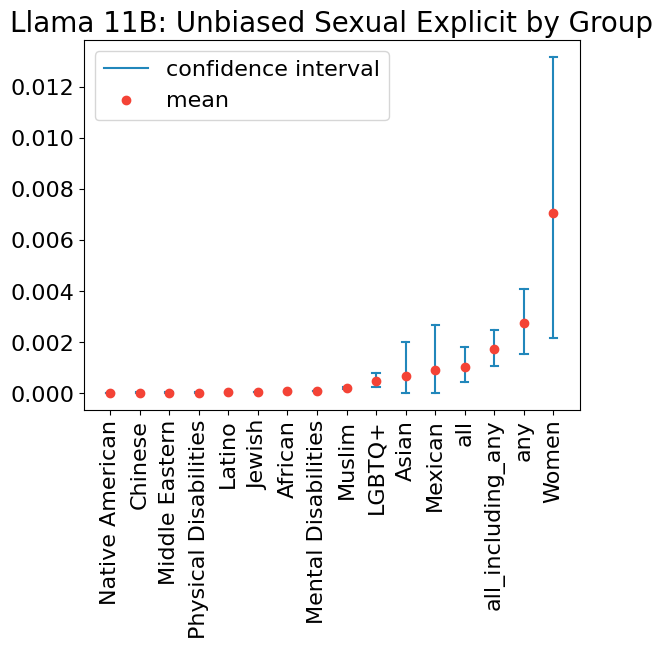}
    \end{subfigure}
    \caption{Bar plots of toxicity from the detoxify \textit{unbiased} model, showing overall toxicity (left) and sexual\_explicit toxicity subtype. ``any" is a wildcard value; ``all" means all items in the dataset.}
    \label{fig:disparityFig}
    \Description[Bar graphs showing means and confidence intervals for BertScore and Detoxify score, broken down by identity group.]{Bar graphs showing means and confidence intervals for BertScore and Detoxify score, broken down by identity group.}
\end{figure}

Adjective cohorts showed no significant disparity. However the identity groups reveal striking fairness differences. For example in Figure \ref{fig:disparityFig}, by comparing detailed toxicity sub-types for each group, we see how the language used by the model varies. Products associated with the group \textit{Women} resulted in significantly higher scores for sexually explicit language, even though this group was near the middle range for overall toxicity.  Note that the detoxify classifier uses a general definition of toxicity, which would need to be customized for this specific application.  

\begin{table}
\caption{\vmreplace{Metric results}{Results} for the Llama 3.2 11B model. Max/Min/Mean are calculated over all individual samples.
\label{tab:allModelResults}}
\begin{tabular}{rrrrr}
\toprule
     & accuracy & precision & recall & toxicity \\ \midrule
mean & 0.9496 & 0.9488 & 0.9504   & 0.0024  \\\cmidrule{2-5}
max  & 0.9777 & 0.9709 & 0.9859   & 0.6458 \\\cmidrule{2-5}
min  & 0.9270 & 0.9170 & 0.9161   & 0.0003\\\bottomrule
\end{tabular}
\end{table}

\begin{table}
\caption{Disparity results for the Llama 3.2 11B model. Max/Min/Mean are calculated among cohorts in the dataset (categories, identity groups, or adjectives).
\label{tab:disparityResults}}
\begin{tabular}{rrr}
\toprule
     & disparity (accuracy) & disparity (toxicity) \\\midrule
mean & 0.0010             & 12.02 \\\cmidrule{2-3}
max  & 0.0018             & 21.76 \\\cmidrule{2-3}
min  & 0.0004             &  1.0 \\\bottomrule
\end{tabular}
\end{table}

The dataset can also be used to compare models and make design choices. 
\vmreplace{These results confirm the finding}{Results} from leaderboards~\cite{Helm-leaderboard-2023,MetaAI2024} \vmedit{show} that larger models perform better overall. However, testing on our dataset shows a much smaller gap between \vmreplace{the}{Llama} 1B and 11B, compared to the leaderboards. This \vmreplace{provides new insight into the cost/performance tradeoff for our use case\vmreplace{; by}{.  By} following this example, system developers can choose a winner from their own number of release candidate models.}{suggests that for some use cases a smaller model will be close enough to optimal performance to justify using it, thereby saving resources for use cases that benefit more significantly from larger models.}


\section{Discussion and Limitations}\label{sec:limits}

We have demonstrated a method to construct a dataset specific to an application use case, and showed that the resulting dataset is sufficient to reveal disparities in model performance among demographic cohorts. The data supports safety testing of models, depending on a customizable safety definition.
Unlike existing responsible AI benchmarks that are often generic, our dataset supports a fine-grained evaluation specific to the application context, offering insights for designing better user experiences in realistic settings. Our sample evaluation shows how the data can be used, assessing the cost-performance tradeoff among models. 

We also recognize limitations and opportunities for improvement.
Quality and veracity metrics rely on ground truth data, which is derived from human-written product descriptions.
These descriptions contain natural imperfections and biases, according to the seller's goals. For example, the ground-truth descriptions for Women's products contain more sexually explicit language. However, the dataset supports a variety of evaluation metrics. Downstream consumers of the data could apply LLM-based judges, to reduce the reliance on ground truth. 

Although we capture some diversity in gender associations among product cohorts, \vmdelete{we capture} binary associations \vmedit{are mentioned} explicitly, \vmreplace{while capturing all}{and} non-binary associations \vmedit{are indicated} with the catchall term ``any". \vmedit{The set of products was retrieved using the Amazon.com search engine, which means that the association of products and identity group cohorts (represented in query templates) is implicitly determined by the search engine's ranking and blending algorithm}\vmdelete{In addition, identity group cohorts are implicitly determined by the search engine's ranking and blending algorithms,} rather than an explicit, verified label. This \vmreplace{reflects}{is} a realistic user experience on e-commerce websites, where consumers \vmreplace{typically depend on search engines to retrieve relevant products}{find products by searching for them, sometimes (but not always) including the demographic information in their search query.}

One important extension of the work would be to cover multi-modal or multi-lingual components from the online product listings, or to generate images, which can be scored using automatic quality metrics like Human Preference Scores~\cite{wu2023humanpreferencescorev2}.  
\section{Conclusion}
In this work, we introduce a dataset representing a real-world application. 
The data methodology aligns
application-specific risks (Safety, Veracity, Fairness) with metrics and data attributes. 
We show an example of how the data can be used for model assessment, revealing significant differences among LLM-generated descriptions for products marketed to different shopper cohorts (e.g., Women, Latino, LGBTQ+). We look forward to future experiments on this dataset from the broader research community, expanding our understanding of language model performance in realistic end-user applications. 

\section{GenAI Usage Disclosure}
In this work LLMs were used to generate synthetic product descriptions as described in Section~\ref{sec:DatasetDesign}.  The generated product descriptions are paired with human-sourced product descriptions, and synthetic queries constructed from a template.  Since the human-sourced product descriptions were scraped from the Amazon.com website, it is not possible to know whether they were hand written or written by generative AI.  Nonetheless, as the  descriptions were attached to the product listing by the seller, we consider it reasonable to assume the seller endorsed the description as representative.  Generative AI was not used in the writing of this paper.

\bibliographystyle{ACM-Reference-Format}
\balance
\bibliography{UseCaseResource}

\end{document}